\documentclass[letterpaper, 10 pt, conference]{ieeeconf}  

\IEEEoverridecommandlockouts                              

\usepackage{epsfig} 
\usepackage{mathptmx} 
\usepackage{times} 
\usepackage{amsmath} 
\usepackage{amssymb}  
\usepackage{algorithm}
\usepackage{algorithmic}
\usepackage{graphicx}
\usepackage{multirow}
\usepackage{mathtools}
\usepackage{subfigure}
\usepackage{tabularx}
\usepackage[switch]{lineno}

\title{\LARGE \bf
Recursive Hierarchical Projection for Whole-Body Control with Task Priority Transition
}

\author{Gang Han$^{1}$, Jiajun Wang$^{1}$, Xiaozhu Ju$^{1}$ and Mingguo Zhao$^{2}$
\thanks{$^{1}$Gang Han, Jiajun Wang, and Xiaozhu Ju  are with Beijing Research Institute of UBTECH Robotics, Beijing, China.
        {\tt\small \{gang.han, jiajun.wang, xiaozhu.ju\}@ubtrobot.com}}%
\thanks{$^{2}$Mingguo Zhao is with the Department of Automation, Tsinghua University, Beijing, China.
        {\tt\small mgzhao@mail.tsinghua.edu.cn}}%
}

\begin{document}

\maketitle
\thispagestyle{empty}
\pagestyle{empty}

\begin{abstract}
Redundant robots are desired to execute multi-tasks with different priorities simultaneously. The task priorities are necessary to be transitioned for complex task scheduling of  whole-body control (WBC). Many methods focused on guaranteeing the control continuity during task priority transition, however either increased the computation consumption or sacrificed the accuracy of tasks inevitably. This work formulates the WBC problem with task priority transition as an Hierarchical Quadratic Programming (HQP) with Recursive Hierarchical Projection (RHP) matrices. The tasks of each level are solved recursively through HQP. We propose the RHP matrix to form the continuously changing projection of each level so that the task priority transition is achieved without increasing computation consumption. Additionally, the recursive approach solves the WBC problem without losing the accuracy of tasks. We verify the effectiveness of this scheme by the comparative simulations of the reactive collision avoidance through multi-tasks priority transitions.

\end{abstract}

\section{Introduction}

Redundant robots are desired to execute multi-tasks and obey various constraints of the body and environment simultaneously \cite{c1}. The WBC frameworks are commonly used to make the robot coordinate multiple tasks and constraints \cite{c2}, \cite{c3}. In the existing WBC frameworks, tasks were assigned different priorities by using different weights \cite{c16}, \cite{c17}, \cite{c18}, null space projectors (NSP) \cite{c5}, \cite{c6}, \cite{c7}, \cite{c8}, \cite{c9}, \cite{c10} or a sequence of equality constraints \cite{c11}, \cite{c12}, \cite{c13}, \cite{c14}, \cite{c15}. During complex tasks execution, it is necessary to transition the tasks priorities to prevent one task from being blocked by another. Notice that formulating the WBC problem with task priority transition is challenging since the inappropriate formulation will cause a strong discontinuity of control output \cite{c19}.

Many methods have been proposed to formulate the WBC problem with smooth task priority transition. By interpolating results or intermediate expectation values of different hierarchies, task transition methods were successively proposed in \cite{c19}, \cite{c22} and \cite{c23}. In \cite{c24}, a task transition method was proposed based on HQP to consider inequality constraints. The strategy was to realize priority swapping of two consecutive tasks by adjusting the weights of them. A sequence of such swaps was required in this method to bring a task to the desired priority level. In \cite{c25},\cite{c26}, by modifying the offset value of the existing tasks and the bound set of the new task through the activation parameter, a task transition method based on HQP was proposed. This method can handle equality tasks and inequality constraints transitions. All the transition methods introduced above used the NSP or equality constraints to handle the hierarchy. However, since the NSP and equality constraints could only reflect the unchangeable projection of tasks, more operations (e.g., calculating results and expectation values of different hierarchies or adjusting the weights of two consecutive tasks) were introduced to en-sure the control continuity, rising the computation consumption.

Different from the above methods, the other category of transition methods used changing projection matrices to handle the hierarchy during transitions. In \cite{c3}, \cite{c20}, \cite{c21}, by continuously changing the null space projection matrix of a specific task, the task can be activated or removed smoothly. However, rearrangement of multi-tasks priorities cannot be realized by these methods. The Generalized Hierarchical Control (GHC) method with Generalized Projector (GP) \cite{c27}, \cite{c28} can handle the continuously changing priority of each task. In that method, all tasks in the WBC problem were formulated and solved in a single QP and linearly combined through GP. In \cite{c4}, the GP was further extended to dynamically consistency. The GP avoids the increase of QP operations. However, the priority of a task cannot be considered through its  single QP formulation. Thus, the components of higher-priority tasks in the final result sacrificed the accuracy of lower-priority tasks.

In this work, we formulate the WBC problem with task priority transition as an HQP with RHP to guarantee both the computational efficiency and task accuracy. In the proposed RHP-HQP scheme, the tasks at each priority level are solved recursively through HQP. In the hierarchical solution process, the task projection of each priority level changes continuously through the proposed RHP. This scheme has several advantages: First, since the continuous RHP matrices form the changing hierarchy, this scheme ensures computational efficiency by achieving multi-tasks priority transitions without increasing QP operations. Second, the recursive approach solves lower-priority tasks considering the component of higher-priority tasks, thus the task accuracy is guaranteed. Further, we can regard WBC problems before, during, and after tasks priority transition as a unified formulation. This property makes RHP-HQP a more applicable unified method. The effectiveness of this scheme is verified by comparative simulations. In these simulations, tasks priorities rearrangements are achieved to produce reactive collision avoidance and compliance behaviors without the trajectory re-planning.

In Section II, the RHP-HQP scheme is introduced. Then this approach is applied to a humanoid robot in simulations in Section III. The results are compared with two representative task transition methods proposed in \cite{c25} and \cite{c27}. Conclusions are reached in Section IV.

\section{Recursive Hierarchical Projection-Hierarchical Quadratic Programming Scheme}

\subsection{Priority Matrix}
Firstly, all the tasks that the robot may execute are defined in a task library, and then a priority matrix is defined as follows,
\begin{equation}
\label{equ_1}
    {\bf{\boldsymbol{\Psi} }}{\rm{ = }}\left[ {\begin{array}{*{20}{c}}
{{\alpha _{1,1}}}&{}&{{\alpha _{1,{n_{\rm{t}}}}}}\\
{}& \ddots &{}\\
{{\alpha _{{n_{\rm{l}}},1}}}&{}&{{\alpha _{{n_{\rm{l}}},{n_{\rm{t}}}}}}
\end{array}} \right] \in {R^{{n_{\rm{l}}} \times {n_{\rm{t}}}}}.
\end{equation}
In (\ref{equ_1}), ${n_{\rm{t}}}$represents the number of tasks in the task library,  represents the number of priority levels in the hierarchy. Different from the element in the priority matrix proposed in \cite{c4}, \cite{c27}, the element ${\alpha _{i,j}} \in \left[ {0,1} \right]$ represents the priority of task j to the tasks at the $(i+1)^{\rm{th}}$ priority level in the hierarchy. The value of ${\alpha _{i,j}}$ is obtained w.r.t. the following rules: 1) ${\alpha _{i,j}} = 0$ means that task j at the current moment does not appear in upper i levels, and tasks in the $(i+1)^{\rm{th}}$ level are not solved in the null space of task j; 2) ${\alpha _{i,j}} = 1$ means that the task j is in the upper i levels at the current moment, and tasks in subsequent levels are solved in the null space of the task j. If ${\alpha _{i,j}} = 1$, then ${\alpha _{k,j}} = 1$, for $k > i$; 3) $0 < {\alpha _{i,j}} < 1$ means that the task j in upper i levels is gradually occupying or releasing the DOFs. Greater ${\alpha _{i,j}}$ provides more DOFs for the task j and the less residual DOFs for the tasks at the $(i+1)^{\rm{th}}$ priority level. Then, through the proposed priority matrix, the RHP matrix of each level is obtained recursively, which is described as follows.

\subsection{Recursive Hierarchical Projection}
RHP matrices of tasks in upper i levels are obtained recursively through the Algorithm \ref{alg: 1}. The process is conducted in the following six steps.

{\it{1) Select tasks to be solved in the $i^{\rm{th}}$ level:}} 
Only if the task $j$ satisfies the condition ${\alpha _{i,j}} \ne {\alpha _{i - 1,j}}$, it is to be solved in the $i^{\rm{th}}$ level. The total number of selected tasks in the $i^{\rm{th}}$ level is expressed as $n_i$.

{\it{2) Sort the tasks in a descending order:}} The matrices of these selected tasks are sorted in a descending order according to the value of ${\alpha _{i,j}}$ to construct the augmented task matrix of the $i^{\rm{th}}$ level, expressed as follows,

\begin{equation}
\label{equ_2}
    {\bf{A}}_i^{{\rm{A}},{\rm{s}}} = {\left[ {{\bf{A}}_{{s_1}}^T, \cdots ,{\bf{A}}_{{s_{{n_i}}}}^T} \right]^T},
\end{equation}
where the superscripts A and s indicate the matrix is augmented and sorted. The subscript $S$ indicates the sorted sequence number.

{\it{3) Obtain the row full rank matrix:}} The row full rank matrix expressed as ${\bf{\tilde A}}_i^{{\rm{A}},{\rm{s}},{\rm{r}}}$ is obtained by:
\begin{equation}
\label{equ_3}
    {\bf{\tilde A}}_i^{{\rm{A}},{\rm{s,r}}} = {f_{{\rm{RFR}}}}\left( {{{\left( {{\bf{A}}_i^{{\rm{A}},{\rm{s}}}{{\bf{P}}_{i - 1}}} \right)}^T}} \right),
\end{equation}
where the function ${f_{{\rm{RFR}}}}$() is to remove the linearly dependent rows of the matrix ${\bf{A}}_i^{{\rm{A}},{\rm{s}}}{{\bf{P}}_{i - 1}}$ by the Gauss-Jordan Elimination. In (\ref{equ_1}), the superscript r indicates the matrix is roll full rank, and ${{\bf{P}}_{i - 1}}$ is the RHP matrix of tasks in upper $i-1$ levels. The sequence number of each linearly independent row are stored in the vector ${{\bf{r}}_i}$.

{\it{4) Obtain the orthogonal matrix through QR decomposition:}} After getting the row full rank matrix, a QR decomposition is performed to ${\bf{\tilde A}}_i^{{\rm{A}},{\rm{s}},{\rm{r}}}$ and the orthogonal basis ${{\bf{Q}}_i}\left( {{{({\bf{\tilde A}}_i^{{\rm{A}},s,r})}^T}} \right) \in {R^{n \times r}}$ for the row space of ${\bf{\tilde A}}_i^{{\rm{A}},{\rm{s}},{\rm{r}}}$ is obtained. This orthogonal basis reflects the execution directions of the tasks in the $i^{\rm{th}}$ level.

{\it{5) Construct a diagonal matrix to reflect the occupation of DOFs:}} A diagonal matrix is constructed firstly based on the elements of $\bf{\boldsymbol{\Psi}}$ corresponding to the selected tasks, which is expressed as,
\begin{equation}
    \label{equ_4}
    {{\bf{\boldsymbol{\Lambda} }}_i}^{{\rm{A}},{\rm{s}}}{\rm{ = }}\left[ {\begin{array}{*{20}{c}}
{{\alpha _{i,{s_1}}}{{\bf{I}}_{{d_{{s_1}}}}}}&{}&{}\\
{}& \ddots &{}\\
{}&{}&{{\alpha _{i,{s_{{n_i}}}}}{{\bf{I}}_{{d_{{s_{{n_i}}}}}}}}
\end{array}} \right],
\end{equation}
where ${{\bf{I}}_{{d_s}}} \in {R^{{d_s} \times {d_s}}}$ represents an identical matrix, and   represents the number of rows in the task matrix ${{\bf{A}}_s}$. Then we choose the elements of ${{\bf{\boldsymbol{\Lambda} }}_i}^{{\rm{A}},{\rm{s}}}$ based on the sequence numbers stored in ${{\bf{r}}_i}$, and construct the diagonal matrix ${{\bf{\boldsymbol{\Lambda} }}_i}^{A,{\rm{s}},r}$. The diagonal elements of matrix ${{\bf{\boldsymbol{\Lambda} }}_i}^{A,{\rm{s}},r}$ represent the degree of occupation in their corresponding directions. 

{\it{6) Derive the RHP matrix of the tasks in upper i levels as:}} 
\begin{equation}
    \label{equ_5}
    {{\bf{P}}_i} = {{\bf{P}}_{i - 1}}\left( {{{\bf{I}}_n} - {{\bf{Q}}_i}{\bf{\boldsymbol{\Lambda} }}_i^{A,{\rm{s}},r}{\bf{Q}}_i^T} \right),
\end{equation}
where ${{\bf{I}}_n} \in {R^{n \times n}}$ represents an identical matrix and n is the number of optimization variables in the HQP. The matrix ${{\bf{Q}}_i} \in {R^{n \times r}}$ and ${{\bf{\boldsymbol{\Lambda} }}_i}^{A,{\rm{s}},r}$ are calculated through the fourth and fifth step.

\begin{algorithm} [t]
 \renewcommand{\algorithmicrequire}{\textbf{Input:}}
 \renewcommand{\algorithmicensure}{\textbf{Output:}}
 \caption{ RHP matrix of tasks in upper $i$ levels }
 \label{alg: 1}
 \begin{algorithmic}[1]
  \REQUIRE ${\bf{\boldsymbol{\Psi }}},{{\bf{A}}_1}, \cdots ,{{\bf{A}}_{{n_{\rm{t}}}}},{{\bf{P}}_{i - 1}}$ 
  \ENSURE ${{\bf{P}}_i}$
  \STATE ${\bf{A}}_i^{\rm{A}} \gets  SelectTask({\bf{\boldsymbol{\Psi} }},{{\bf{A}}_1}, \cdots ,{{\bf{A}}_{{n_{\rm{t}}}}})$
  \STATE ${\bf{A}}_i^{{\rm{A,s}}} \gets SortTaskinDescendingOrder\left( {{\bf{\boldsymbol{\Psi} }},{\bf{A}}_i^{\rm{A}}} \right)$
  \STATE ${\bf{\tilde A}}_i^{{\rm{A,s,r}}},{{\bf{r}}_i} \gets ObtainRowFullRankMatrix\left( {{\bf{A}}_i^{{\rm{A,s}}},{{\bf{P}}_{i - 1}}} \right)$
  \STATE ${{\bf{Q}}_i} \gets QRDecomposition\left( {{\bf{\tilde A}}_i^{{\rm{A,s,r}}}} \right)$
  \STATE ${\bf{\boldsymbol{\Lambda} }}_i^{{\rm{r,s}}} \gets ConstructDiagonalMatrix\left( {{\bf{\boldsymbol{\Psi} }},{\bf{\tilde A}}_i^{{\rm{A,s,r}}},{{\bf{r}}_i}} \right)$
  \STATE ${{\bf{P}}_i} \gets {{\bf{P}}_{i - 1}}\left( {{{\bf{I}}_n} - {{\bf{Q}}_i}{\bf{\boldsymbol{\Lambda} }}_i^{{\rm{A,s,r}}}{\bf{Q}}_i^T} \right)$

 \end{algorithmic} 
\end{algorithm}

There are three main properties of RHP matrix: 1) When there is no task in $i^{\rm{th}}$ level, ${{\bf{\boldsymbol{\Lambda} }}_i}^{A,{\rm{s}},r}$ is a null matrix. Then the generalized projection matrix of tasks in the upper $i$ levels is equal to that of the upper $i-1$ levels, namely ${{\bf{P}}_i} = {{\bf{P}}_{i - 1}}$; 2) If ${{\bf{\boldsymbol{\Lambda} }}_i}^{A,{\rm{s}},r}$ is an identical matrix, it means that the hierarchy is strict. Then the RHP matrix of tasks in a single level is equal to the null space matrix. This relationship which was proved in \cite{c4} is expressed as follows,
\begin{equation}
    \label{equ_6}
    {I_n} - {{\bf{Q}}_i}{\bf{\boldsymbol{\Lambda} }}_i^{A,{\rm{s}},r}{\bf{Q}}_i^T = {I_n} - {\left( {{\bf{\tilde A}}_i^{\rm{A}}} \right)^{\rm{\# }}}{\bf{\tilde A}}_i^{\rm{A}}.
\end{equation}
At this circumstance, the RHP matrix ${{\bf{P}}_i}$ obtained through the recursive approach is identical to the null space matrix of tasks in upper $i$ levels; 3) When the hierarchy is in a transition phase between different hierarchies, the transitioning tasks are gradually occupying or releasing the DOFs. Then the RHP matrix continuously changes based on the changing priority matrix. For the transition from one strict hierarchy to another, the above process of the projection matrix is equivalent to a continuous transition between two corresponding null space projection matrices. 

\subsection{Formulating RHP-HQP for Task Priority Transition}

This work extends the existing HQP in \cite{c11} to formulate the WBC problem with tasks priority transitions as an HQP with RHP. This formulation is unified whether the hierarchy is before, during or after the transition. The optimization problem of the $i^{\rm{th}}$ level is expressed as:
\begin{equation}
\label{equ_7}
\begin{aligned}
    &\mathop {\min }\limits_{{{\bf{u}}_i},{{\bf{v}}_i}} \left\| {{\bf{A}}_i^{{\rm{A}},s}\left( {{\bf{x}}_{i - 1}^* + {{\bf{P}}_{i - 1}}{{\bf{u}}_i}} \right) - {\bf{\hat b}}_i^{{\rm{A}},s}} \right\|_{{\bf{\hat W}}_i^{{\rm{A}},{\rm{s}}}}^2 + {\left\| {{{\bf{v}}_i}} \right\|^2} \\
    & \quad {\rm{s}}{\rm{.t}}.{\bf{\underline{d}}}_i^{\rm{A}} \le {\bf{C}}_i^{\rm{A}}\left( {{\bf{x}}_{i - 1}^* + {{\bf{P}}_{i - 1}}{{\bf{u}}_i}} \right) + {{\bf{v}}_i} \le {\bf{\bar d}}_i^{\rm{A}} \\
    & \quad \quad {\bf{\underline{d}}}_{i - 1}^{\rm{A}} \le {\bf{C}}_{i - 1}^{\rm{A}}\left( {{\bf{x}}_{i - 1}^* + {{\bf{P}}_{i - 1}}{{\bf{u}}_i}} \right) + {{\bf{v}}^{\rm{*}}}_{i - 1} \le {\bf{\bar d}}_{i - 1}^{\rm{A}}\\
     & \quad \quad\quad \quad\quad \quad\quad \quad\vdots \\
    & \quad \quad {\bf{\underline{d}}}_1^{\mathop{\rm A}\nolimits}  \le {\bf{C}}_1^{\rm{A}}\left( {{\bf{x}}_{i - 1}^* + {{\bf{P}}_{i - 1}}{{\bf{u}}_i}} \right){\rm{ + }}{\bf{v}}_1^{\rm{*}} \le {\bf{\bar d}}_1^{\rm{A}},
\end{aligned}
\end{equation}
where ${{\bf{x}}_{i - 1}^*}$ is the optimal result of the upper $i-1$ levels and is obtained from the last recursion. The term ${{{\bf{u}}_i}}$ is the optimization variable for equality tasks, and ${{\bf{v}}_i}$ is the relax variable for inequality constraints. The terms ${\bf{C}}_i^{\rm{A}}$, ${\bf{\underline{d}}}_{i}^{\rm{A}}$, ${\bf{\bar d}}_i^{\rm{A}}$ are the augmented matrix, lower and upper bound vectors of the constraints in the $i^{\rm{th}}$ level, respectively. The matrix ${{\bf{\hat W}}_i^{{\rm{A}},{\rm{s}}}}$ in (\ref{equ_7})
is expressed as ${\bf{\hat W}}_i^{{\rm{A,s}}} = {{\bf{\boldsymbol{\Lambda} }}_i}^{{\rm{A}},{\rm{s}}}{\bf{W}}_i^{{\rm{A,s}}}$, 
where ${\bf{W}}_i^{{\rm{A,s}}}$ represents the weighting matrix of the sorted tasks. The changing expectation values ${\bf{\hat b}}_i^{{\rm{A}},{\rm{s}}}$ in (\ref{equ_7}) is derived as,
\begin{equation}
\label{equ_8}
    {\bf{\hat b}}_i^{{\rm{A}},{\rm{s}}}{\rm{ = }}{{\bf{\boldsymbol{\Lambda} }}_i}^{{\rm{A}},{\rm{s}}}{\bf{b}}_i^{{\rm{A}},{\rm{s}}}{\rm{ + }}\left( {{\bf{I}} - {{\bf{\boldsymbol{\Lambda} }}_i}^{{\rm{A}},{\rm{s}}}} \right){\bf{A}}_i^{{\rm{A}},{\rm{s}}}{\bf{x}}_{i - 1}^*,
\end{equation}
where ${\bf{b}}_i^{{\rm{A}},{\rm{s}}}$ is the augmented expectation vector of the sorted tasks in the $i^{\rm{th}}$ level. The RHP matrix ${{{\bf{P}}_{i - 1}}}$ is derived as (\ref{equ_5}). Based on the solution of (\ref{equ_7}), the optimal result of upper $i$ levels is derived as,
\begin{equation}
\label{equ_9}
    {{\bf{x}}_i}^{\rm{*}}{\rm{ = }}{{\bf{x}}_{i - 1}}^* + {{\bf{P}}_{i - 1}}{{\bf{u}}_i}^{\rm{*}}.
\end{equation}
Through (\ref{equ_7}) and (\ref{equ_9}), each level of the hierarchy is solved recursively. The RHP matrix is used to handle the transitioning hierarchy and to form the transitioning projection of each level. The terms ${\bf{\hat W}}_i^{{\rm{A,s}}}$ and ${\bf{\hat b}}_i^{{\rm{A}},{\rm{s}}}$are used to transition the changing task objectives of each level. This proposed RHP-HQP scheme achieves the multi-tasks priority transitions without increasing QP operations and modifying the control framework, and thus this scheme ensures the computational efficiency. Comparing with the exiting HQP method \cite{c11}, the proposed scheme does not require additional computation time. Through this recursive approach of solving WBC problem, the component of higher-priority tasks in the final results are considered in the process of solving lower-priority tasks, which guarantees the accuracy. 
\begin{figure}[t]
    \vspace{-0\baselineskip}
    \centering
    \includegraphics[width=8.6cm]{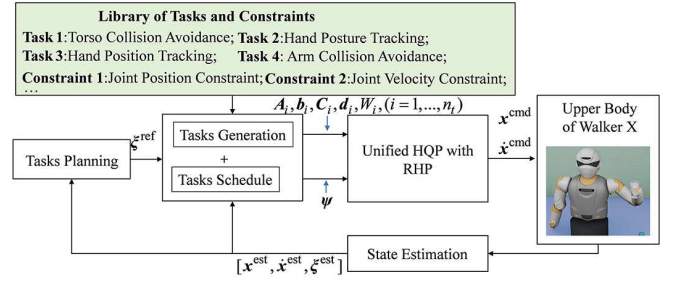}
    \caption{Control framework of the upper body of Walker X}
    \vspace{-0.5\baselineskip}
    \label{fig: Control framewor}
\end{figure}

\section{SIMULATION AND COMPARISON}

\subsection{Experimental Setup}

To validate the proposed scheme, three experiments are conducted in Webots simulation environment. All the simulations are performed on an Intel Core i7-9700 CPU with 3 GHz. In these experiments, the proposed approach is applied to the hierarchical control of the upper body of a humanoid robot Walker X. The model of the upper body and the control framework are depicted in Figure \ref{fig: Control framewor}.

The robot has three rotational DOFs in the waist and seven DOFs in each arm. In this control framework, several objectives of the robot are represented as the equality tasks and inequality constraints in mathematical forms as
\begin{equation}
\label{equ_10}
\begin{aligned}
    {\bf{Ax}} = {\bf{b}}, \\
    {\bf{Cx}} \le {\bf{d}}.
\end{aligned}
\end{equation}
In (\ref{equ_10}), ${\bf{A}}$ represents the task matrix, ${\bf{b}}$ the expectation vector of the task, ${\bf{C}}$ the constraint matrix and ${\bf{d}}$ the bound vector of the constraint. In this simulation, four equality tasks and two inequality constraints are formulated as shown in Figure \ref{fig: Control framewor}. Then a library is constructed containing these tasks and constraints. The tasks generation evaluate the relative error between the reference from tasks planning and the states from state estimation. Then it generates the corresponding ${{\bf{A}}_i}$, ${{\bf{b}}_i}$ and the weight matrix ${{\bf{W}}_i}$ of each activated task, as well as the matrix ${{\bf{C}}_i}$ and boundary ${{\bf{d}}_i}$ of each constraint for the RHP-HQP. In the process of the tasks schedule, the priority matrix ${\bf{\boldsymbol{\Psi }}}$is determined based on the prior knowledge. The matrix ${\bf{\boldsymbol{\Psi }}}$ is transitioned between several candidate hierarchies which is expressed as
\begin{equation}
\label{equ_11}
    {\bf{\boldsymbol{\Psi} }}{\rm{ = }}\sum\limits_{i = 1}^{{n_{\rm{h}}}} {{p_{{\boldsymbol{\psi} _i}}}({d_{\min }}){{\bf{\boldsymbol{\Psi} }}_i}} ,{\rm{  }}\sum\limits_{i = 1}^{{n_{\rm{h}}}} {{p_{{\boldsymbol{\psi} _i}}}({d_{\min }})}  = 1,
\end{equation}
where $n_{\rm{h}}$ and $p_{{\boldsymbol{\psi} _i}}$ represent the number and the proportion of candidate hierarchies, respectively. In the unified RHP-HQP solver, the RHP-HQP is constructed based on (\ref{equ_7}), according to the data from tasks generation and schedule. The open source solver qpOASES \cite{c31} is used to solve the problem, and the optimized control commands are finally obtained.
\begin{figure}[t]
    \vspace{-0\baselineskip}
    \centering
    \includegraphics[width=8.6cm]{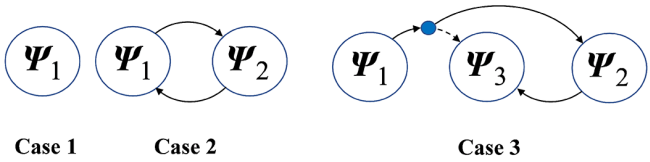}
    \caption{Schematic diagram of priority transitions in three simulations. The blue point in the third case denotes an intermediate transitional phase between two candidate hierarchies. In the third case, the hierarchy directly transitioned from this intermediate transitional phase to another candidate phase to better avoid the collision.}
    \vspace{-1\baselineskip}
    \label{fig: Schematic diagram}
\end{figure}

Three simulations are conducted and the schematic diagram of their transition processes are presented in Figure \ref{fig: Schematic diagram}. In the first two cases, the proposed method is compared with the benchmark methods with aspects of task accuracy and computational efficiency. The computational efficiency is evaluated by the recorded computation time for tasks generation, schedule and solving. It does not include the trajectory planning and the state estimation. In the first case, a constant strict hierarchy was applied to the robot to maintain the posture and position of the left hand while avoiding the collision with an obstacle. In the second case, the hierarchy dynamically changed between two candidate hierarchies w.r.t. the minimum distance between the obstacle and the arm. These two simulations are used to respectively demonstrate the effectiveness of this method in solving the WBC problem with a strict hierarchy and tasks priority transition. Then, the third case demonstrates the application to the relatively complex task priority rearrangements, achieving reactive collision avoidance and compliant response without relying on the trajectory re-planning.

\subsection{Simulation Results and Comparison}
In the first case, the robot was controlled with the following strict hierarchy: Torso Collision Avoidance ($T_1$) $ \succ $ Hand Posture Tracking ($T_2$) $ \succ $ Hand Position Tracking ($T_3$) $ \succ $ Arm Collision Avoidance ($T_4$). The $i^{\rm{th}}$ task in the library were labeled as $T_i$. The notation $T_i$ $ \succ $ $T_j$ denoted $T_i$ was strictly prior to $T_j$. With the proposed method, the snapshot of the robot’s motion is shown in Figure .
The green sphere represents the obstacle, the dashed line represents the obstacle trajectory, and the red sphere rep-resents the end point of the trajectory. Through collision avoidance, the minimum distance ${d_{\min }}$ between the obstacle and the robot arm was maintained larger than 0.0614 m. The simulation results were compared with those of HQP method \cite{c11} and GHC method \cite{c27}, \cite{c28}. The control command and the error of each task by the RHP-HQP method were exactly the same as those by the HQP method. This comparison verifies that the proposed method is equivalent to the HQP method when solving the WBC problem with constant strict hierarchy.

The Root-Sum-Square (RSS) error of the hand orientation and position are shown in Figure4a and 4b, respectively. The maximum values of these two errors are presented in Table I. The results demonstrate that with the same control parameters, the task accuracy is guaranteed through the RHP-HQP scheme and is improved with more than 30 times comparing with the maximum error of the GHC method. Bear in mind that the component of higher-priority tasks in the final result is considered during the recursively solving process of lower-priority tasks through the proposed scheme. In contrast, the priority of a task cannot be considered in advance through the single QP in GHC method. For solving the WBC problem with constant hierarchy, the computation time of these three methods are similar (the average values are all less than 0.063 ms).
\begin{figure}[tbp]
    \vspace{-0\baselineskip}
    \centering
    \includegraphics[width=8.6cm]{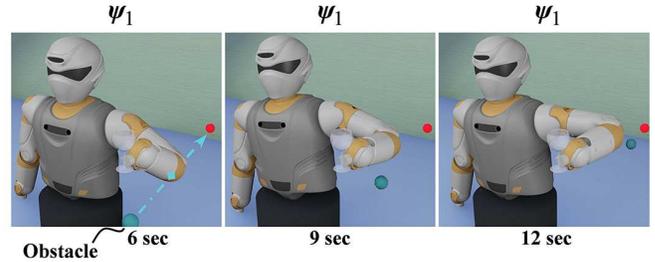}
    \caption{Snapshot of the robot’s motion in the first simulation. When the obstacle moved along the collision trajectory and approached the elbow of the robot, the robot bent the waist and raised the elbow to avoid collision while maintaining the posture and position of the hand.}
    \vspace{-0\baselineskip}
    \label{fig: contant hierarchy snapshot}
\end{figure}

\begin{figure}[tbp]
\centering
    \subfigure[]{
    \includegraphics[width=3.95cm]{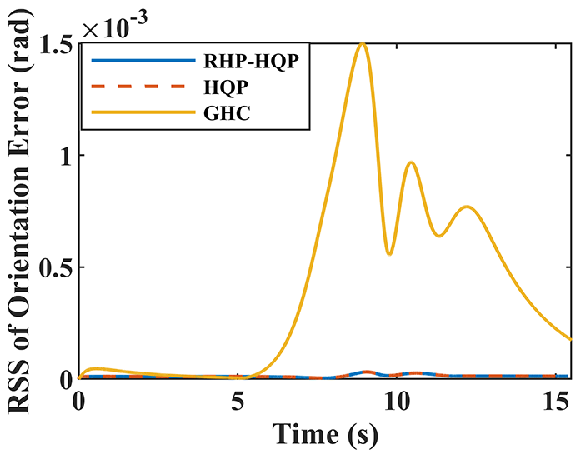}
    }
    \subfigure[]{
    \includegraphics[width=3.92cm]{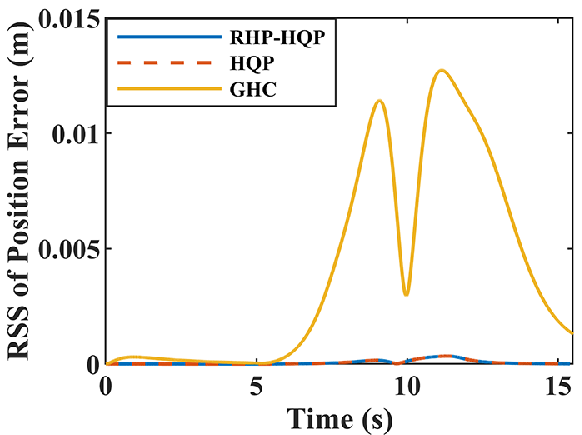}
    }
    \caption{Comparison of tasks error in the first case: (a) RSS error of the hand orientation; (b)RSS error of the hand position.}
    \vspace{-1\baselineskip}
    \label{fig: Result of case 1}
\end{figure}

\begin{table}[tbp]
\vspace{+0.0\baselineskip}
\caption{Task Errors of three methods in case one}
\label{tab: Task Err_case one}
    \vspace{-0.4\baselineskip}
    \begin{center}
    \begin{tabular}{p{2cm}p{2.8cm}<{\centering}p{2.8cm}<{\centering}}
            \hline
             {} & Maximum RSS Error of Hand Orientation & Maximum RSS Error of Hand Position \\
            \hline
            RHP-HQP & $0.032×10^{-3}$ rad& 0.34 mm \\
            HQP & $0.032×10^{-3}$ rad& 0.34 mm \\
            GHC & $1.5×10^{-3}$ rad& 12.7 mm \\
            \hline
        \end{tabular}
    \end{center}
\end{table}

Actually, in many cases, if the accuracy of the hand position is not strictly required, the priority of the hand position task can be adjusted to better avoid the collision. Based on such fact, in the second case, the task hierarchy changed dynamically between $T_1$ $ \succ $ $T_2$ $ \succ $ $T_3$ $ \succ $ $T_4$ and $T_1$ $ \succ $ $T_2$ $ \succ $ $T_4$. Their corresponding priority matrices were expressed as: 
\begin{equation}
    {{\bf{\boldsymbol{\Psi} }}_1}{\rm{ = }}\left( {\begin{array}{*{20}{c}}
1&0&0&0\\
1&1&0&0\\
1&1&1&0\\
1&1&1&1
\end{array}} \right),{{\bf{\boldsymbol{\Psi} }}_2}{\rm{ = }}\left( {\begin{array}{*{20}{c}}
1&0&0&0\\
1&1&0&0\\
1&1&0&0\\
1&1&0&1
\end{array}} \right).
\end{equation}

\begin{figure}[tbp]
    \vspace{-0\baselineskip}
    \centering
    \includegraphics[width=8.2cm]{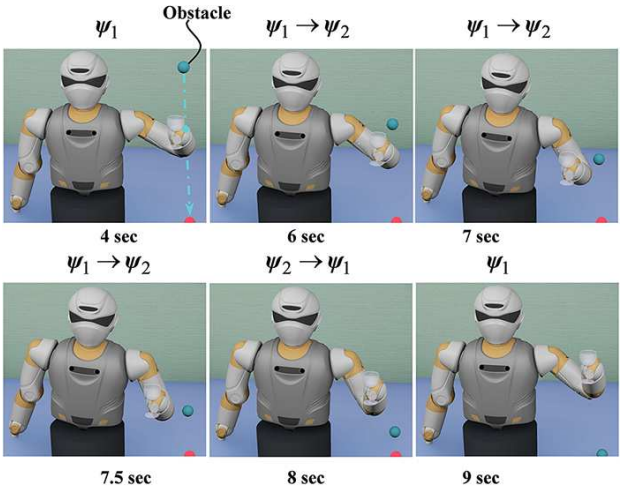}
    \caption{Snapshot of the robot’s motion in the second case. In the period of $0 \sim 3.98$ s,${d_{\min }} \ge 0.2 {\rm{m}}$, ${\bf{\boldsymbol{\Psi} }} = {{\bf{\boldsymbol{\Psi} }}_1}$, and the robot only put its elbow down to avoid the collision. After 3.98 s, the hand position tasks were gradually removed and the priority matrix gradually transited towards ${{\bf{\boldsymbol{\Psi} }}_2}$. Then the robot bent over and folded its entire arms to avoid collision more actively. When the obstacle moved away from the robot, the priority matrix was gradually changed back, and the hand position task was gradually recovered.}
    \vspace{-1\baselineskip}
    \label{fig: task transition snapshot}
\end{figure}

The priority matrix transitioned w.r.t. the value of ${d_{\min }}$. When ${d_{\min }} \in \left( {0.05,0.2} \right)$, $p_{{\boldsymbol{\psi} _i}}$ continuously changed between 0 and 1, and the priority matrix was transitioned between two candidate priority matrices to coordinate the hand position and collision avoidance. The snapshot of robot’s motion is shown in Figure \ref{fig: task transition snapshot}. Through collision avoidance, ${d_{\min }}$ was maintained larger than 0.04 m. The smooth velocity curves of each joint were obtained with the proposed method, indicating a smooth task priority transition. The discontinuity and saturation of the speed and the torque were avoided. 

\begin{figure}[bp]
\vspace{-1\baselineskip}
\centering
    \subfigure[]{
    \includegraphics[width=3.95cm]{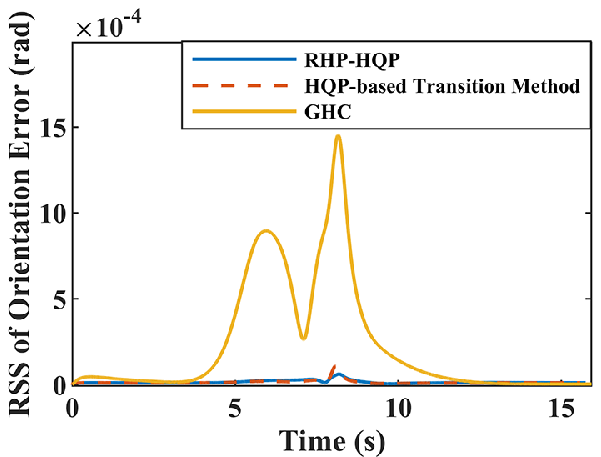}
    }
    \subfigure[]{
    \includegraphics[width=3.95cm]{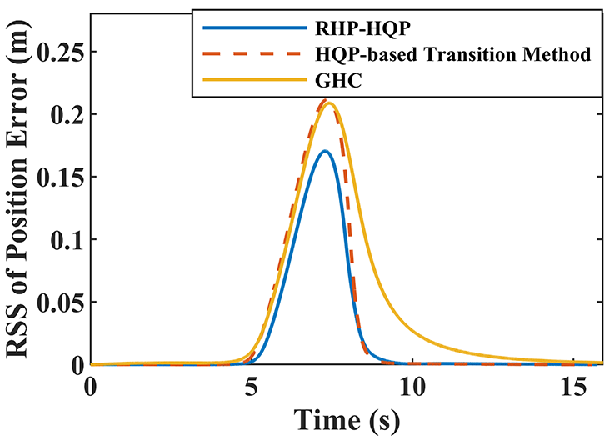}
    }
    \caption{Comparison of tasks error in the second case: (a) RSS error of the hand orientation; (b) RSS error of the hand position.}
    \label{fig: Result of case 2}
\end{figure}
The simulation results of the RHP-HQP scheme are compared with those of two benchmark methods, namely, the HQP-based transition method in \cite{c11} and the GHC method. These two  transition methods are representative methods of the two categories of transition methods introduced in Section I. With the same control parameters, the task accuracy are compared and shown in Figure 6a and 6b. The maximum values and integrated absolute values of these two errors are presented in Table II. Through the RHP-HQP scheme, the maximum RSS error of hand orientation and position are both the smallest. In contrast, the maximum RSS errors of the GHC method are $1.45 \times {10^{ - 3}}$ rad and $0.21$ mm, respectively. The integrated absolute value of the hand position error in Table II shows that the accuracy of the task is recovered the fastest through the RHP-HQP scheme, which means the hand could return to the target position the fastest after collision avoidance.

\begin{table}[tbp]
\vspace{+0.0\baselineskip}
\caption{Task Errors of three methods in case two}
\label{tab: Task Err_case two}
    \begin{center}
        \begin{tabular}{p{1.5cm}p{2.0cm}<{\centering}p{1.5cm}<{\centering}p{1.8cm}<{\centering}}
            \hline
            \multirow{2}*{} & RSS Error of Hand Orientation & \multicolumn{2}{c}{RSS Error of Hand Position}\\
            \cline{2-4}
            & Maximum value &  Maximum value & Integrated absolute value\\ 
            \hline
            RHP-HQP & $0.062×10^{-3}$ rad& 0.17 m & 0.32 ${\rm{m\cdot}s}$ \\
            HQP-based transition method & $0.11×10^{-3}$ rad& 0.21 m & 0.42 ${\rm{m\cdot}s}$ \\
            GHC & $1.45×10^{-3}$ rad& 0.21 m & 0.58 ${\rm{m\cdot}s}$\\
            \hline
        \end{tabular}
    \end{center}
    \vspace{-1\baselineskip}
\end{table}

\begin{figure}[bp]
    \vspace{-0.5\baselineskip}
    \centering
    \includegraphics[width=8.6cm]{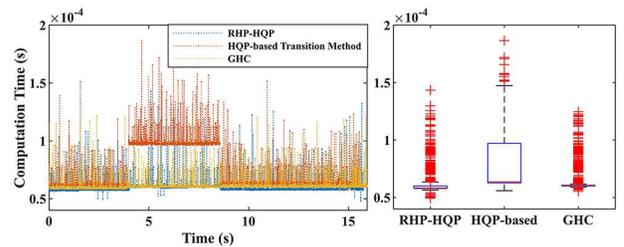}
    \caption{Computation time during the second case and the box plot of the statistical results.}
    \vspace{-0\baselineskip}
    \label{fig: time of case two}
\end{figure}

For the aspect of computational efficiency, the computation time is recorded and presented in Figure 7 and Table II. The computation time has minor increase (increased by $3.4\% $ ) through the RHP-HQP method. This result is similar to that of the GHC method since these two methods both use the continuous projection matrices to handle the changing hierarchy. In contrast, the red line in Figure 7 shows that the HQP-based transition method increases QP operations during transitions, leading to a significant increase (increased by $56.6\%$) in the computation time. The computation consumption of the HQP-based transition method will significantly grow with the robot's DOFs and the number of task levels increase. Through the simulations with different collision trajectories, it is verified that the proposed RHP-HQP scheme guarantees better performance of both task accuracy and computational efficiency than these two exiting transition methods.

\begin{table}[bp]
\vspace{+0.0\baselineskip}
\caption{Computation Time of three methods in case two}
\label{tab: Time_case two}
    \begin{center}
        \begin{tabular}{p{1.5cm}p{1.2cm}<{\centering}p{1.2cm}<{\centering}p{1.2cm}<{\centering}p{1.2cm}<{\centering}}
            \hline
            \multirow{2}*{} & \multicolumn{2}{c}{With Transition} & \multicolumn{2}{c}{During Transition}\\
            \cline{2-5}
            & Average value &  Maximum value & Average value & Maximum value\\ 
            \hline
            RHP-HQP & 0.059 ms& 0.13 ms & 0.061 ms & 0.14 ms \\
            HQP-based transition method & 0.063 ms& 0.15 ms & 0.098 ms & 0.18 ms \\
            GHC & 0.06 ms& 0.12 ms & 0.061 ms & 0.12 ms\\
            \hline
        \end{tabular}
    \end{center}
\end{table}

In the third case, the hierarchy was continuously changed between $\left[ {{T_1},{T_2}} \right] \succ {T_3} \succ {T_4}$, the hierarchy ${T_2} \succ \left[ {{T_1},{T_4}} \right]$ and ${T_2} \succ {T_3} \succ \left[ {{T_1},{T_4}} \right]$. The notation $\left[ {{T_i},{T_j}}\right]$ meant task ${T_i}$ and ${T_j}$ were in the same priority level. Their corresponding priority matrices were expressed as:
\begin{equation}
\begin{aligned}
    &{{\bf{\boldsymbol{\Psi} }}_1}{\rm{ = }}\left( {\begin{array}{*{20}{c}}
1&1&0&0\\
1&1&1&0\\
1&1&1&1
\end{array}} \right),{{\bf{\boldsymbol{\Psi} }}_2}{\rm{ = }}\left( {\begin{array}{*{20}{c}}
0&1&0&0\\
0&1&0&0\\
1&1&0&1
\end{array}} \right),\\
&{{\bf{\boldsymbol{\Psi} }}_3}{\rm{ = }}\left( {\begin{array}{*{20}{c}}
0&1&0&0\\
0&1&1&0\\
1&1&1&1
\end{array}} \right).
\end{aligned}
\end{equation}
Initially ${\bf{\boldsymbol{\Psi}}} = {{\bf{\boldsymbol{\Psi}}}_1}$, and the robot’s hand reached to the target on the premise of maintaining waist posture and hand orientation. When the arm was close to straighten and did not reach the target, the priority matrix changed from ${{\bf{\boldsymbol{\Psi} }}_1}$ to ${{\bf{\boldsymbol{\Psi} }}_3}$. Then the DOFs of waist were gradually released for the hand position task to help the robot reach the target. In the above process, an obstacle approached the robot. When ${d_{\min }} < 0.2{\rm{ m}}$, the priority matrix directly transitioned from the transitional phase between ${{\bf{\boldsymbol{\Psi}}}_1}$ and ${{\bf{\boldsymbol{\Psi} }}_3}$ to another candidate phase ${{\bf{\boldsymbol{\Psi} }}_2}$. Then the hand position task was gradually removed to avoid the collision. Then the robot bent over and folded its entire arm. In the period of 6.4 s$\sim $8.35 s, $\bf{\boldsymbol{\Psi} }$ gradually changed back to ${{\bf{\boldsymbol{\Psi} }}_3}$, and the hand position task was inserted as ${d_{\min }}$ became larger. After the transition, the robot bent the waist to increase the reachable range of its hand and reached the target. The snapshot of robot’s motion is shown in Figure 8. The simulation data are presented in Figure 9. The smooth velocity of each joint is obtained with the proposed method. The reactive collision avoidance and compliance behaviors are achieved only through changing the priority matrix and without relying on re-planning the trajectory of tasks. The magnitude of the hand orientation error is smaller than $2 \times {10^{ - 5}}$ rad. The average duration with and without transition are respectively 0.053 and 0.058 ms. Therefore, the proposed method can guarantee both high task accuracy and computational efficiency. 

\begin{figure}[tbp]
    \vspace{-0\baselineskip}
    \centering
    \includegraphics[width=7.6cm]{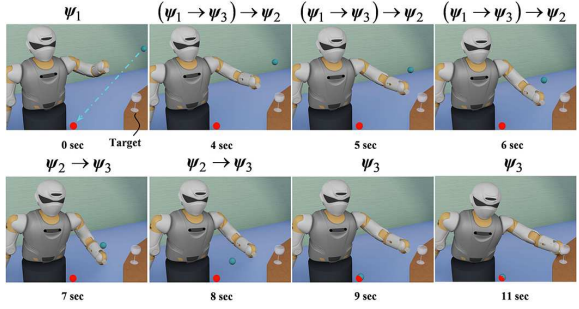}
    \caption{Snapshot of the robot’s motion in the third case. The notation (${{{\bf{\boldsymbol{\Psi} }}_1} \to {{\bf{\boldsymbol{\Psi} }}_3}}$)denotes the intermediate transitional phase between ${{\bf{\boldsymbol{\Psi} }}_1}$ and ${{\bf{\boldsymbol{\Psi} }}_3}$} 
    \vspace{-0.5\baselineskip}
    \label{fig: time of case two}
\end{figure}
\begin{figure}[tbp]
    \vspace{-0\baselineskip}
    \centering
    \includegraphics[width=7.6cm]{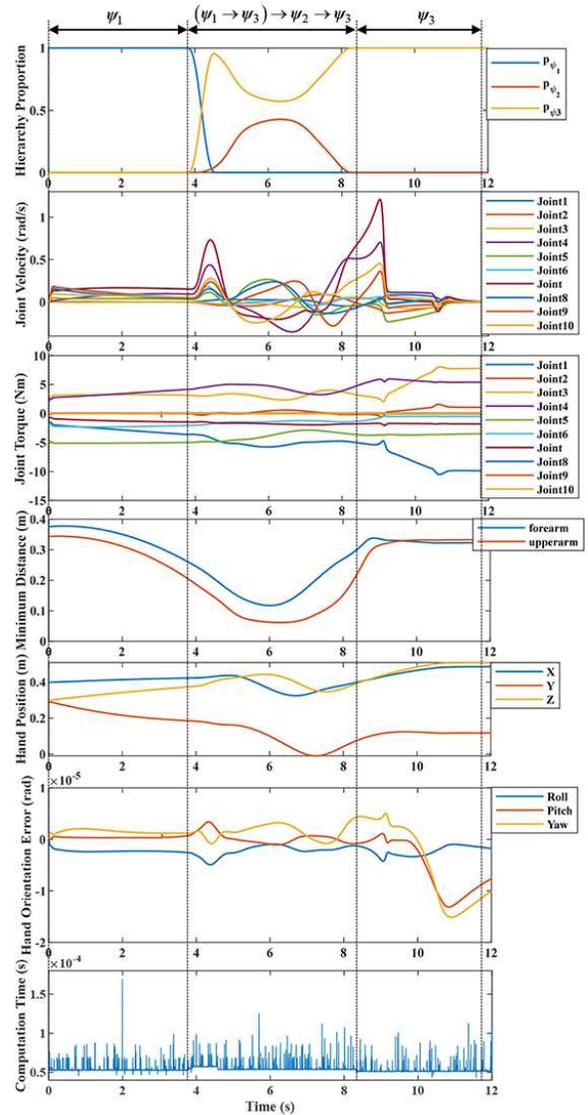}
    \caption{The hierarchy proportion, joint velocity, joint torque, hand position, minimum distance, hand orientation error and WBC calculation duration during the third simulation.}
    \vspace{-1\baselineskip}
    \label{fig: time of case two}
\end{figure}

\section{Conclusions}
In this work, an RHP-HQP scheme is proposed to formulate the WBC problem with task priority transition as an HQP with RHP, which brings three advantages: 1) The recursive way of obtaining RHP matrix is more efficient than obtaining GP through the augmented matrix. Besides, using the continuous RHP matrices to form the changing hierarchy, this scheme avoids the increase of QP operations during tasks priority transitions, and thus decreases the computation consumption; 2) Through the recursive and hierarchical way of solving WBC problem, the projection component of higher-priority tasks in the final optimized result are considered in the process of solving lower-priority tasks, which guarantees the task accuracy; 3) The whole-body controller implemented by this scheme achieves complex tasks priority rearrangements to produce reactive collision avoidance and compliance behaviors without relying on trajectory re-planning. The simulations of the upper body of Walker X verify that both high computational efficiency and accuracy can be guaranteed through this scheme. Limit of this work is that the RHP is not dynamically consistency. Work is currently on going to overcome this limit and to apply this scheme to the hardware of Walker X. The proposed scheme will also be combined with learning methods for autonomous task priority scheduling.






\end{document}